\documentclass[11pt,a4paper]{article}
\usepackage[T1]{fontenc}
\usepackage[utf8]{inputenc}
\usepackage{newtxtext,newtxmath}
\usepackage[margin=2.5cm]{geometry}
\usepackage{graphicx,booktabs,tabularx,array}
\usepackage{cite}
\usepackage[labelsep=period,font=small]{caption}
\usepackage[hidelinks]{hyperref}

\makeatletter\renewcommand\@seccntformat[1]{\csname the#1\endcsname.\quad}\makeatother
\newcolumntype{Y}{>{\raggedright\arraybackslash}X}
\makeatletter
\def\bstctlcite{\@ifnextchar[{\@bstctlcite}{\@bstctlcite[@auxout]}}
\def\@bstctlcite[#1]#2{\@bsphack\@for\@citeb:=#2\do{\edef\@citeb{\expandafter\@firstofone\@citeb}\if@filesw\immediate\write\csname #1\endcsname{\string\citation{\@citeb}}\fi}\@esphack}
\makeatother

\title{Large Language Models: A Plain-Language Overview for Engineers}
\author{Bahaeddin Türkoğlu\\ \small Department of Artificial Intelligence and Data Engineering, Ankara University, Ankara, Türkiye\\ \small \href{mailto:turkoglub@ankara.edu.tr}{\texttt{turkoglub@ankara.edu.tr}} \quad ORCID: \href{https://orcid.org/0000-0003-0255-8422}{0000-0003-0255-8422}}
\date{}

\begin{document}
\bstctlcite{IEEEexample:BSTcontrol}
\maketitle

\begin{abstract}
Large language models (LLMs) are neural networks trained on huge text collections to generate text by predicting what comes next. Engineers using ChatGPT and other generative artificial intelligence (AI) tools need a clear picture of what training changes and prompts add. The article starts with the Transformer, which connects words in context, and pre-training, where models learn to predict text. It then explains how fine-tuning adapts models to requests, with reinforcement learning from human feedback using human answer rankings. Next, retrieval-augmented generation (RAG) supplies documents, while chain-of-thought prompting encourages intermediate steps and agents use tools. FlashAttention lowers training and running costs by reducing data movement during attention; quantized low-rank adaptation (QLoRA) saves fine-tuning memory. Readers get a life cycle figure and tables comparing methods and their limits. A short checklist for judging LLM evaluation claims flags test material in training (data contamination) and model-based grading (LLM-as-a-judge). It also covers very long inputs (long context) and hallucination, meaning false or unsupported claims. Check factual accuracy separately from whether an answer sounds helpful.
\end{abstract}

\noindent\textbf{\textit{Index Terms}}---Foundation models, Generative artificial intelligence, ChatGPT, Transformer, Reinforcement learning from human feedback, Retrieval-augmented generation, Chain-of-thought prompting, LLM agents, Hallucination, Language model evaluation.

\section{Introduction}\label{sec:1}

An engineer pastes a short function into ChatGPT and gets a fluent explanation back. The hard part is deciding how far to trust it. ChatGPT gives engineers a conversational interface to large language models (LLMs), one type of generative artificial intelligence (AI). A large language model is a neural network trained on large text collections to generate text by predicting what comes next. But the chat box alone does not explain how these generative AI tools produce answers. Knowing how LLMs work, and what the surrounding system adds, helps engineers decide what to test before using an answer.

Training and prompt choices matter because they can change the answers a model produces. Brown et al. tested GPT-3 using instructions and examples in the prompt, without further task-specific training. Their evaluation specified different tasks through text rather than changing the model for each task. But the authors left open whether examples taught unfamiliar tasks or helped the model recognize tasks encountered during training \cite{brown2020language}. Ouyang et al.’s labelers preferred answers from a smaller InstructGPT model, trained with human feedback, to those from a much larger GPT-3 model \cite{ouyang2022training}. In that study, greater model size did not lead to preferred answers; the comparison did not establish usefulness on engineering workloads.

This article gives engineers and researchers outside natural language processing a short map of the LLM life cycle. It separates pre-training on large text collections from post-training, which further trains a model to shape its behavior. Inference-time augmentation adds methods while the model answers a request, rather than changing it through further training. For example, retrieval-augmented generation (RAG) supplies retrieved documents to help a model produce an answer. Evaluation and safety checks apply across these stages, while efficient infrastructure reduces the work or memory needed to train and run models. Keeping these parts separate helps readers judge whether a reported gain applies to their own workload.

Section II explains the Transformer architecture and language model training; Section III examines alignment, shaping behavior toward stated human goals, and its measurement limits. Section IV covers in-context learning, using prompt examples without further training, and scaling laws relating model size, training data and computation. Post-training follows in Section V, including fine-tuning, which adapts a trained model through further training. Section VI covers chain-of-thought prompting, which supplies intermediate reasoning steps, alongside RAG and agents that choose actions with tools. Section VII addresses efficient attention and memory-saving fine-tuning methods, while Section VIII gives a checklist for checking whether tests support reported comparisons. Section IX examines hallucination, generating false information or claims unsupported by supplied evidence, alongside other open problems.

\section{The Transformer and language model training}\label{sec:2}

The Transformer is a neural network architecture that uses attention to combine information from different positions in a sequence. It provides the architecture for LLMs such as GPT-3. Vaswani et al. introduced an encoder to read input and a decoder to produce output. Both parts alternate attention layers with feed-forward networks, which process each position on its own \cite{vaswani2017attention}. The architecture separates how positions exchange information from how each position processes it.

Self-attention lets each token, a word or part of a word, gather information from other tokens in the text. The model scores how relevant those tokens are, then gives more weight to information with higher scores. For example, nearby words can help distinguish a river bank from a bank that handles payments. The word “river” supplies context for interpreting “bank”. Attention combines that context with information from the other words \cite{vaswani2017attention}.

Attention heads let the model track different relationships within the same text. Each head makes its own set of relevance scores rather than sharing a single set. Vaswani et al. reported that individual heads appeared to track aspects of sentence structure and meaning. The model also adds position codes, numerical signals that mark where tokens occur in the sequence \cite{vaswani2017attention}. These codes supply order information alongside the token information.

Decoder-only language models generate text by predicting the next token from the text already available. Brown et al. used this version of the Transformer for GPT-3 \cite{brown2020language}. During pre-training, the model learns from a large text collection by repeatedly making these predictions. Training adjusts its parameters, the learned numerical weights, to reduce prediction error. During generation, the model adds a token and uses the extended text to predict the next one.

\begin{figure}[!t]
\centering
\includegraphics[width=\linewidth]{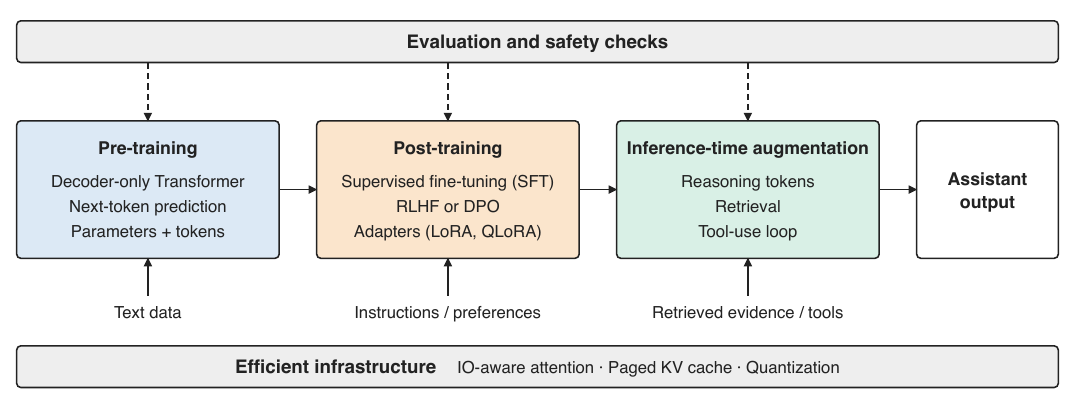}
\caption{The life cycle of a large language model. Pre-training builds a base model from large text collections; post-training adapts it to follow instructions and preferences; inference-time augmentation adds reasoning, retrieval and tool use. Evaluation and safety checks apply at every stage, and efficient infrastructure underlies all of them.}\label{fig:1}
\end{figure}

Fig.~\ref{fig:1} shows three life-cycle stages: pre-training, post-training, and inference-time augmentation. Evaluation and safety checks extend across all stages, with efficient infrastructure supporting them underneath. This separates changes made during training from methods used while the trained model answers requests.

Pre-training builds the base model through learning from text, as described above and developed further in Section~\ref{sec:4}. It supplies the starting model for post-training. Post-training changes that model to help it follow instructions and human preferences \cite{ouyang2022training}. Section~\ref{sec:5} covers those training methods. Inference-time augmentation adds methods used while a trained model handles a request, including outside information or tool use \cite{lewis2020retrievalaugmented,yao2023react}. Section~\ref{sec:6} covers these methods.

Evaluation and safety checks are continuing work, not a final stage after training. Section~\ref{sec:3} explains why stronger test scores do not settle questions about behavior or safety. Section~\ref{sec:8} covers assessment, while Section~\ref{sec:7} explains infrastructure methods that reduce the work or memory needed to train and run models.

\section{Alignment and measurement for language models}\label{sec:3}

A model that predicts text well is not automatically helpful, truthful or safe. Alignment is the process of shaping a model’s behavior to match stated human goals and constraints. But matching human preferences does not establish truthfulness or safety \cite{ouyang2022training}.

More training data does not ensure that more groups are fairly represented. Bender et al. \cite{bender2021on} argued that web participation, data collection and filtering favor already dominant views. They warned that generated text could repeat or amplify biases that harm marginalized groups. Their argument puts data choice alongside data size as an engineering concern.

Human feedback can help a smaller model give answers that people prefer. In Ouyang et al.’s labeler study, labelers preferred the 1.3-billion-parameter InstructGPT to the 175-billion-parameter GPT-3. They also rated InstructGPT better at following stated constraints and answering as a customer assistant \cite{ouyang2022training}.

Better instruction following still leaves room for hallucination: generating information that is false or unsupported by supplied evidence. In Ouyang et al.’s summaries and question-answering tasks, hallucination specifically meant adding information absent from the supplied text. InstructGPT’s hallucination rate was 21\%, compared with GPT-3’s 41\% \cite{ouyang2022training}.

Safety results depended on the prompt and the test. On RealToxicityPrompts, a test of toxic text generation, InstructGPT produced fewer toxic outputs when asked to be respectful. Its advantage over GPT-3 disappeared without that request. It also showed no significant improvement on Winogender and CrowS-Pairs, tests of gender and social bias \cite{ouyang2022training}.

Human feedback reflects the people who give it. Ouyang et al. used about 40 mostly English-speaking labelers and almost entirely English training data \cite{ouyang2022training}. The authors said this group did not represent the full range of people who would use or be affected by their models. Users in other languages or communities may prefer different answers.

Preference scores and knowledge tests can move apart. Zheng et al. improved GPT-4-rated scores on MT-Bench, a multi-turn conversation test, by training on selected conversations. But scores on Massive Multitask Language Understanding (MMLU), a multiple-choice knowledge test, changed little \cite{zheng2023judging}. So engineers should test factual accuracy separately from conversational preference.

\section{In-context learning and scaling laws}\label{sec:4}

In-context learning means performing a task using instructions or examples in the input, without changing the model’s weights. Brown et al. tested GPT-3 this way, without task-specific fine-tuning \cite{brown2020language}. Their few-shot learning prompts supplied worked examples before the input the model had to answer. These prompts typically contained 10 to 100 examples. This let the researchers test different tasks through text prompts rather than separate training runs.

Prompt examples improved GPT-3’s answers on TriviaQA, a question-answering test. Accuracy rose from 64.3\% without examples to 71.2\% with few-shot examples \cite{brown2020language}. The model answered in a closed-book setting, without consulting outside documents. So this comparison measured how examples helped it answer using information already available in the model. Across the authors’ tests, the benefit of examples often grew with model size.

Few-shot learning did not work equally well across tasks. GPT-3 struggled with some reading-comprehension tasks and comparisons of word meanings \cite{brown2020language}. For example, it did little better than chance when deciding whether words had the same meaning in different sentences. Brown et al. also left open whether prompt examples taught unfamiliar tasks or helped GPT-3 recognize tasks encountered during training. Better scores alone could not settle that distinction.

Pre-training data choices determine which text enters the base model’s training set. Brown et al. used filtered Common Crawl web text mixed with curated collections \cite{brown2020language}. That mixture makes data selection part of the engineering work, rather than just a matter of collecting more text. Bender et al. recommended including documentation in the planned costs of dataset creation \cite{bender2021on}. They advised collecting only as much data as teams could thoroughly document within that budget.

Beyond choosing the training text, teams must divide a fixed training budget between model size and data. Training compute is the arithmetic used to train a model; scaling laws relate that budget to model size, data and prediction error. Hoffmann et al. fitted these relationships using more than 400 trained models \cite{hoffmann2022training}. Their fits favored growing model size and training data in about equal proportions as compute increased. Earlier fits by Kaplan et al. had favored model size more strongly \cite{kaplan2020scaling}. A larger model can leave too little budget for training it on enough text.

Chinchilla tested whether spending more of the budget on data could beat a larger model. It was four times smaller than Gopher and trained on more text using the same training compute \cite{hoffmann2022training}. Its smaller size also reduced memory needs and the cost of running it. This comparison held the training budget fixed, rather than giving the better-scoring model more computation.

On MMLU, Chinchilla’s average five-shot accuracy was 67.6\%, compared with Gopher’s 60.0\% \cite{hoffmann2022training}. But Chinchilla and Gopher were the study’s only comparable large-scale training runs. The authors had no additional tests at intermediate scales. They also warned that extrapolation to much larger computing budgets remained uncertain, so the fitted balance was not a guarantee at every scale.

\section{Post-training and fine-tuning methods}\label{sec:5}

Instruction tuning trains a language model on instructions paired with example answers so it learns how to respond to requests. Unlike the in-context learning in Section~\ref{sec:4}, instruction tuning changes the model’s weights. Preference training adds a different signal: which answer a person likes better. These methods differ in their data needs and training work. Table~\ref{tab:I} compares the four methods by the supervision they need, their training work and their main limitation.

\begin{table}[!t]
\caption{Post-training methods: the supervision they need, the training work and the main limitation stated in each paper.}\label{tab:I}
\small
\begin{tabularx}{\linewidth}{>{\hsize=0.850\hsize}Y>{\hsize=0.850\hsize}Y>{\hsize=1.049\hsize}Y>{\hsize=1.251\hsize}Y}
\toprule
\textbf{Method} & \textbf{Supervision} & \textbf{Training work} & \textbf{Main limitation stated in the papers} \\
\midrule
Supervised instruction tuning \cite{ouyang2022training} & Human-labeler supervision & Supervised fine-tuning & Labelers are not representative; data is almost entirely English \\
RLHF \cite{ouyang2022training} & Human comparisons of model outputs & Fit preference-predicting reward model; optimize outputs using reinforcement learning & Narrow labeler population; benchmark regressions persist despite mixing pretraining updates \\
DPO \cite{rafailov2023direct} & Pairs of preferred and dispreferred responses & Direct preference updates; no separate reward model or reinforcement-learning loop & Larger-scale testing remains open; generalization evidence is preliminary \\
QLoRA \cite{dettmers2023qlora} & OASST1 and subsampled FLAN datasets in chatbot experiments & Adapter fine-tuning across all linear transformer layers & Full fine-tuning parity unestablished at larger evaluated scales \\
\bottomrule
\end{tabularx}
\end{table}

Supervised fine-tuning (SFT) needs examples that show what a good response looks like. Ouyang et al. used labeler-written answers to begin training InstructGPT. The model learned to produce those answers from their prompts, without a separate model to score responses. But the source of those examples mattered. On prompts from their service’s users, GPT-3 trained on public instruction datasets did worse than their own supervised model \cite{ouyang2022training}.

Human rankings provide a way to train beyond those example answers. Reinforcement learning from human feedback (RLHF) uses human rankings of answers to guide further training. In Ouyang et al.’s pipeline, a reward model learned to predict labelers’ preferences. The language model then generated answers and received scores from that reward model. Proximal policy optimization (PPO), a reinforcement learning algorithm, updated the language model’s weights using those scores. This adds reward-model training and repeated answer generation to SFT \cite{ouyang2022training}.

That added computation was still small compared with pre-training. For their largest models, Ouyang et al. reported 4.9 petaflop/s-days for SFT and 60 for PPO with pre-training updates. GPT-3’s pre-training used 3,640 in the same units \cite{ouyang2022training}. Petaflop/s-days measure total computation by combining processing rate with time. These figures compare training compute, not the cost of collecting human labels. So they do not describe the whole cost of post-training.

Preference training can lower scores on other tasks, an effect called an alignment tax. Ouyang et al. found that PPO training lowered scores on some public language-processing tests. Mixing pre-training updates into preference training reduced these losses without lowering labeler preference scores. But some losses remained, including those on translation \cite{ouyang2022training}. The safety limits discussed in Section~\ref{sec:3} also remained; preference training did not remove them.

Direct preference optimization (DPO) trains on preferred and rejected answers without a separate reward model or reinforcement learning loop. Rafailov et al.’s training rule favors preferred answers while limiting changes from a fixed reference model. That reference provides a baseline for training, rather than another model that scores generated answers. DPO does not generate fresh answers during fine-tuning, which removes work required by PPO. It still needs a preferred and a rejected answer for each training prompt \cite{rafailov2023direct}.

DPO’s simpler training also performed better in the authors’ TL;DR, or “too long; didn’t read,” summarization comparison. GPT-4 judged DPO’s win rate at about 61\%, compared with 57\% for PPO. Both results used a sampling temperature of zero, giving deterministic answer generation. But the authors tested models only up to six billion parameters. They also said generalization beyond the training distribution needed more study, and judge prompts affected win rates \cite{rafailov2023direct}.

Memory-efficient fine-tuning addresses a different constraint from preference training. Quantized low-rank adaptation (QLoRA) is a fine-tuning method designed to reduce memory use. Section~\ref{sec:7} explains its memory mechanics and reported savings. Dettmers et al. did not establish that QLoRA matched full 16-bit fine-tuning at the 33- and 65-billion-parameter scales \cite{dettmers2023qlora}. Lower memory needs therefore did not establish equivalent quality at those scales.

Data choice still mattered when using QLoRA. A 9k-sample conversation dataset outperformed a 450k-sample public instruction-data subset on the authors’ chatbot benchmarks. That comparison favored the smaller dataset, not simply more examples. Their knowledge-test and chatbot scores also did not consistently move together. So neither measurement could replace the other \cite{dettmers2023qlora}.

\section{Reasoning, retrieval and agents}\label{sec:6}

\begin{figure}[!t]
\centering
\includegraphics[width=\linewidth]{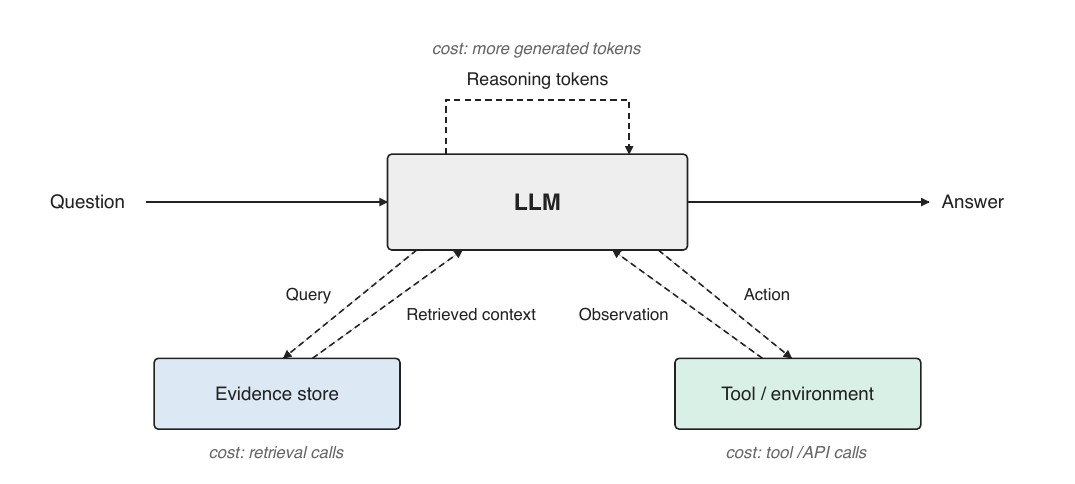}
\caption{Three ways to add capability at inference time without changing model weights: a reasoning loop that writes intermediate steps, retrieval from an evidence store, and actions on tools or an environment. Each adds its own cost in tokens, latency or external calls.}\label{fig:2}
\end{figure}

Unlike the fine-tuning methods above, chain-of-thought prompting supplies worked reasoning examples without changing the model’s weights. These examples encourage the model to generate intermediate steps before answering \cite{wei2022chainofthought}. Fig.~\ref{fig:2} compares reasoning, retrieval and tool use, including their costs.

Worked reasoning examples helped most on problems requiring several steps. Wei et al.’s 540-billion-parameter PaLM model improved on GSM8K, a math word-problem benchmark, from 17.9\% accuracy with standard prompting to 56.9\%. But on SingleOp, a benchmark subset requiring one arithmetic step, gains were small or negative. In their experiments, gains appeared only at about 100 billion parameters. The authors identified this large-model requirement as a serving cost \cite{wei2022chainofthought}.

A readable explanation is not proof that an answer is correct. Wei et al. found errors in generated chains and left open whether they showed that the network was actually reasoning \cite{wei2022chainofthought}. An explanation can help someone inspect an answer, but it does not independently verify it. So teams should test accuracy on their own tasks, including cases where intermediate steps look convincing.

Post-training can also teach LLMs to generate intermediate steps without worked examples in the prompt. Guo et al. trained DeepSeek-R1-Zero using reinforcement learning (RL), which learns through rewards for generated responses. Their rule-based rewards checked answer accuracy and output format, without initial supervised fine-tuning. On the American Invitational Mathematics Examination (AIME) 2024 benchmark, single-answer accuracy rose from 15.6\% to 77.9\% during training \cite{guo2025deepseekr}. That result measures competition mathematics, not general engineering reliability.

DeepSeek-R1’s longer reasoning shows the trade-off between answer quality and generated text. This separate model used supervised fine-tuning as well as RL. On the authors’ 2024 competition-math collection, it averaged 8,793 thinking tokens and solved 61.8\% of problems. GPT-4o averaged 711 output tokens and solved 24.7\% on the same collection. The authors also observed excessive reasoning on simpler questions. They identified scaling RL to tasks without reliable evaluation as an open challenge \cite{guo2025deepseekr}.

Retrieval-augmented generation (RAG) is a method that uses retrieved documents to help a language model answer a question. It supplies evidence beyond information stored in the model’s weights. Lewis et al. jointly fine-tuned the retriever’s query encoder and answer generator, keeping the document encoder and index fixed. In a small world-leader test, the 2016 Wikipedia index gave 70\% accuracy for 2016 leaders but only 4\% for 2018 leaders. Swapping the index changed answers without retraining the generator \cite{lewis2020retrievalaugmented}.

Retrieval makes document selection and placement part of the reliability problem. In Lost in the Middle, Liu et al. found lower accuracy when relevant evidence appeared in the middle of the input \cite{liu2024lost}. So finding the right document does not ensure the model uses it well. Lewis et al. also warned that external sources can contain errors and bias \cite{lewis2020retrievalaugmented}. Retrieved documents are evidence to inspect, not proof of correctness.

An LLM agent can choose actions, such as calling a search tool, and use results to decide its next step. Retrieved information can then guide further actions, rather than just support a final answer. Yao et al.’s ReAct, which combines reasoning and acting, alternates reasoning with actions and observations. On ALFWorld, an interactive task benchmark, its best trial succeeded 71\% of the time, against 45\% for the best action-only trial \cite{yao2023react}.

Agents depend on useful feedback and can get stuck. In Yao et al.’s HotpotQA failure analysis, uninformative search accounted for 23\% of ReAct errors. The authors also reported reasoning errors and repetitive loops \cite{yao2023react}. As deployment guidance, these failures support limiting repeated actions and checking tool results before consequential decisions.

\section{FlashAttention and memory-efficient fine-tuning}\label{sec:7}

Running these systems efficiently requires attention to data movement as well as arithmetic. FlashAttention computes exact attention while moving less data between a graphics processing unit’s (GPU’s) main memory and fast on-chip memory \cite{dao2022flashattention}. Data movement affects runtime, while memory capacity determines what fits. Table~\ref{tab:II} compares four ways to cut cost across the life cycle, from pre-training to preference tuning.

\begin{table}[!t]
\caption{Four ways to cut the cost of large language models across the life cycle: what each saves, a reported result and the main caveat.}\label{tab:II}
\small
\begin{tabularx}{\linewidth}{>{\hsize=0.850\hsize}Y>{\hsize=0.850\hsize}Y>{\hsize=0.850\hsize}Y>{\hsize=1.487\hsize}Y>{\hsize=0.963\hsize}Y}
\toprule
\textbf{Method} & \textbf{Stage} & \textbf{What it saves} & \textbf{Reported result} & \textbf{Caveat} \\
\midrule
Compute-optimal scaling (Chinchilla) \cite{hoffmann2022training} & Pre-training & Model size for a fixed compute budget & Same training compute as Gopher with a model four times smaller, so memory and inference cost drop & Only two large comparable runs; no tests at intermediate scales \\
FlashAttention \cite{dao2022flashattention} & Training and inference & Reads and writes to GPU main memory & GPT-2 medium attention: 40.3 GB of memory traffic cut to 4.4 GB; 41.7 ms cut to 7.3 ms & Each attention variant needs a hand-written GPU kernel; analysis covers one GPU \\
QLoRA \cite{dettmers2023qlora} & Fine-tuning & GPU memory for weights and optimizer state & 65-billion-parameter model fine-tuned in under 48 GB; 4-bit with double quantization matched 16-bit on MMLU & Match with full 16-bit fine-tuning not shown at 33B and 65B \\
DPO \cite{rafailov2023direct} & Preference tuning & A separate reward model and the reinforcement-learning loop & Win rate about 61\% vs 57\% for PPO on summarization, judged by GPT-4 & Tested only up to 6B parameters \\
\bottomrule
\end{tabularx}
\end{table}

FlashAttention changes how attention runs, rather than approximating the calculation. Standard attention stores a large matrix of scores linking text positions. Dao et al. divide the work into blocks that fit in on-chip memory, avoiding storage of the full matrix. During training, they recompute that matrix in the backward pass, which calculates the gradients used to update weights. Their input/output (IO)-aware design accounts for data movement, rather than counting arithmetic alone \cite{dao2022flashattention}.

Moving less data can save time even when the calculation requires more arithmetic. In Dao et al.’s GPT-2 medium attention test, main-memory reads and writes fell from 40.3 to 4.4 gigabytes. FlashAttention ran faster despite performing more arithmetic \cite{dao2022flashattention}. So arithmetic counts alone would give a misleading picture of this operation’s runtime.

The speed gains also extended to complete training runs, not just attention calculations. On OpenWebText, GPT-2 small trained up to three times faster than the HuggingFace baseline at the same prediction quality. But each new attention variant required a hand-written GPU program, with uncertain transfer across GPU architectures. The authors’ data-movement analysis also covered only a single GPU, not transfers between GPUs \cite{dao2022flashattention}.

QLoRA’s memory savings, introduced in Section~\ref{sec:5}, start with how it stores the base model. Quantization stores numbers with fewer bits, reducing the space needed for model weights. QLoRA keeps those base weights fixed in a 4-bit format and trains small adapters, added sets of trainable weights. Dettmers et al. fine-tuned a model with 65 billion parameters using less than 48 gigabytes of GPU memory \cite{dettmers2023qlora}.

QLoRA also reduces the memory needed for information that accompanies the quantized weights. Double quantization compresses the constants used to encode those weights. This saves memory beyond the reduction from storing the base weights with fewer bits \cite{dettmers2023qlora}. The memory accounting therefore includes these constants, rather than treating compressed weights as the only storage cost.

Paged optimizers address another part of QLoRA’s training memory use. These optimizers, the routines that update trainable weights, matched regular optimizers’ training speed at batch size 16 in the reported test. But the authors left measurements of paging-related slowdowns for future work \cite{dettmers2023qlora}. That result does not guarantee unchanged training speed in other conditions. The quality limits remain as described in Section~\ref{sec:5}.

\section{How to assess LLM evaluation claims}\label{sec:8}

LLM evaluation measures how well models perform specific tasks under stated test conditions. First, ask whether the benchmark, a shared test for comparing systems, matches the intended job. Small score gaps also need uncertainty estimates before they support a ranking. Jain et al. estimated about 1 to 1.5\% variation in LiveCodeBench code-generation performance from its limited test size \cite{jain2025livecodebench}.

Data contamination occurs when test material appears in training, so high scores can reflect memory rather than skill. Jain et al. built LiveCodeBench from contest problems with known publication dates. Teams can therefore select problems published after a model's training cutoff, the date its training data ends. DeepSeek models and GPT-4o showed sharp drops on LeetCode problems after release or cutoff dates, suggesting contamination \cite{jain2025livecodebench}. Request those dates, but remember that contest problems do not represent all programming work.

The grader matters as much as the test. LLM-as-a-judge uses one language model to grade another model's answers. Zheng et al. found that GPT-4 agreed with human experts on MT-Bench in 85\% of comparisons without ties. Agreement fell toward 70\% for models close in quality, and judges could favor answers shown first or longer answers \cite{zheng2023judging}. Swap answer order, check a sample with humans, and report the judge prompt.

Long-context tests should move the evidence as well as change input length. A context window limits how much text a model accepts, without guaranteeing reliable use. In Liu et al.'s multi-document question-answering test, GPT-3.5-Turbo's performance could drop by more than 20\% as the answer position changed \cite{liu2024lost}. Test evidence positions and input lengths with distractions resembling the intended workload.

Grounding, meaning support for an answer's factual claims in supplied evidence, needs a separate test from fluency. On Natural Questions, Lewis et al.'s model sometimes answered correctly even when no retrieved document contained the answer \cite{lewis2020retrievalaugmented}. So correctness alone does not establish document support. Inspect sources for each claim, and test behavior when evidence is missing or wrong.

Inference budgets, the computation allowed when producing answers, must be stated for fair comparisons. Guo et al. raised DeepSeek-R1-Zero's AIME 2024 score from 77.9\% to 86.7\% using majority voting over sampled answers \cite{guo2025deepseekr}. The score therefore depends on the answer-generation procedure, not just the model. Report generated tokens and tool calls alongside accuracy and response time.

Table~\ref{tab:III} collects these checks for assessing evaluation claims.

\begin{table}[!t]
\caption{A checklist for reading an LLM evaluation claim, with the papers that motivate each check.}\label{tab:III}
\small
\begin{tabularx}{\linewidth}{>{\hsize=0.850\hsize}Y>{\hsize=1.039\hsize}Y>{\hsize=0.973\hsize}Y>{\hsize=1.137\hsize}Y}
\toprule
\textbf{Check} & \textbf{Evidence to request} & \textbf{What remains unproven} & \textbf{Motivating finding} \\
\midrule
Test contamination \cite{guo2025deepseekr,jain2025livecodebench} & Benchmark version, problem dates, training cutoff, and post-cutoff scores & Post-cutoff subsets can be noisy; text-overlap filters cannot exclude paraphrased tests & Some tested models scored sharply lower on problems released after their cutoffs \\
LLM-as-judge validity \cite{zheng2023judging,rafailov2023direct,dettmers2023qlora} & Answer-order swaps, length-bias tests, and calibration against human votes & Human agreement does not guarantee correctness; judge results depend on prompts & Repetitive verbosity fooled tested judges, with substantially different failure rates \\
Long-context use \cite{liu2024lost} & Accuracy across evidence positions and context lengths, not merely window size & Generality beyond the tested question-answering and synthetic retrieval tasks remains untested & Extended windows did not improve accuracy when both versions accommodated the input \\
Factual grounding \cite{lewis2020retrievalaugmented} & Verify that retrieved passages support each answer claim; assess source quality & Retrieved sources may contain factual errors and bias & Mismatching index dates and question dates reduced accuracy in the world-leader test \\
Inference budget and reasoning \cite{yao2023react,wei2022chainofthought,guo2025deepseekr} & Request token counts, sample counts, relevant tool calls, latency, and accuracy & Plausible reasoning traces do not guarantee correct reasoning & Competition-math evaluation reported thinking-token usage; simpler questions can trigger overthinking \\
Alignment and impact scope \cite{ouyang2022training,bender2021on} & Evaluator population, task coverage, compute, training-data documentation, and environmental disclosures & Narrow truthfulness tests and incomplete risk audits cannot establish broad safety & InstructGPT used a nonrepresentative labeler population and remained incompletely aligned and unsafe \\
\bottomrule
\end{tabularx}
\end{table}

\section{Open problems in large language models}\label{sec:9}

Hallucination remains an open problem for LLMs, even when users supply the evidence an answer should use. Hallucination means generating false information or claims that the evidence given to the model does not support. On tasks requiring answers drawn only from supplied text, Ouyang et al. found fewer unsupported additions with InstructGPT than with GPT-3 \cite{ouyang2022training}. But users still have to check whether sources support a fluent answer. Researchers should test tools that check each claim against its source. Models should also flag missing evidence rather than fill gaps with unsupported claims.

Alignment, which aims to make models follow human goals and preferences, does not settle whose preferences count when people disagree. Ouyang et al. used a small, mostly English-speaking labeler group that they described as unrepresentative of everyone affected \cite{ouyang2022training}. Their results cannot settle what other communities would want from the same system. This matters because people who never rate answers can still be affected by them. Training methods should involve more communities and record disagreements rather than hide them.

Evaluation needs to distinguish test scores from performance in real use. Contamination, where test questions enter training data, makes it hard to separate memorized answers from skill on unfamiliar problems \cite{jain2025livecodebench}. Zheng et al. studied LLM judges, which use a language model such as GPT-4 to grade other models' answers. They noted that their study emphasized helpfulness and largely neglected safety \cite{zheng2023judging}. So preference for an answer alone does not establish that a system is safe to deploy. Researchers should use new problems and real workloads, with separate safety checks.

Written reasoning raises questions about both the steps and their cost. Chain-of-thought prompting asks models to write intermediate steps, but Wei et al. left open whether the model is actually reasoning \cite{wei2022chainofthought}. Guo et al. reported that DeepSeek-R1 sometimes overthinks simpler questions, generating unnecessarily long reasoning \cite{guo2025deepseekr}. The practical question is when extra text improves accuracy enough to justify the work spent producing it. Experiments could edit or delete written steps, then measure changes in answers and generation cost.

Environmental cost extends beyond the money spent on training. Bender et al. cited an estimate of 284 tonnes of carbon dioxide from training a large Transformer with neural architecture search \cite{bender2021on}. That search tests model designs automatically. The estimate applies to that training procedure, not every language model or chat request. The authors also recommend budgeting for data curation and documentation. Teams should report emissions on a comparable basis and fund documentation as part of dataset creation.

\section{Conclusion}\label{sec:10}

In generative AI applications, a large language model system combines a trained model with the software and hardware used to produce answers. Pre-training data and scale shape the base model \cite{hoffmann2022training}; post-training improves instruction following using human feedback \cite{ouyang2022training}. Inference-time augmentation changes how models answer, for example by supplying retrieved documents \cite{lewis2020retrievalaugmented} or prompting intermediate steps \cite{wei2022chainofthought}. Efficient implementations reduce training time \cite{dao2022flashattention} or fine-tuning memory \cite{dettmers2023qlora}. So engineers need to assess the system they will actually use.

For engineering use, measurement is the weakest link: tests capture only part of what matters. An answer people prefer can still contain made-up facts \cite{ouyang2022training}. And the helpfulness scores studied by Zheng et al. largely leave safety unmeasured \cite{zheng2023judging}. Contamination, when test questions enter training data, can inflate reported scores \cite{jain2025livecodebench}. So interpreting scores requires clear test conditions and tests that match the intended job.

For deployment, start by checking how reported results were measured, including the test data and grading method. Test factual accuracy and safety separately on your own workload. When comparing systems on accuracy and response time, also track generated text and tool calls. Trust should depend on those workload-specific results, rather than a benchmark ranking alone.

\section*{Acknowledgment}
Generative artificial intelligence tools helped draft and revise text under the author's direction, produce layout drafts for Figs.~\ref{fig:1}--\ref{fig:2} that were redrawn as vector graphics, and cross-check claims against cited papers; the author checked all content and takes responsibility.

\bibliographystyle{IEEEtran}
\bibliography{references}
\end{document}